\documentclass[man, floatsintext]{apa7}

\usepackage{lipsum}
\usepackage[natbibapa]{apacite}
\usepackage[american]{babel}
\usepackage{dirtytalk}
\usepackage{amsmath}
\usepackage{tcolorbox}
\usepackage{makecell}
\usepackage{setspace}
\definecolor{mpigreen}{HTML}{007977}
\DeclareMathOperator*{\argmax}{arg\,max}

\usepackage{amsthm}
\usepackage{hyperref}

\usepackage{tikz}
\usetikzlibrary{shapes,arrows}

\definecolor{yucky}{HTML}{808000}
\definecolor{yuckx}{HTML}{a03623}
\definecolor{mpigreen}{HTML}{007977}
\definecolor{BrickRed}{HTML}{B6321C}
\definecolor{ForestGreen}{HTML}{009B55}
\usetikzlibrary{arrows.meta}
\definecolor{pyplot0}{HTML}{1f77b4}
\definecolor{pyplot1}{HTML}{ff7f0e}
\definecolor{pyplot2}{HTML}{2ca02c}
\definecolor{pyplot3}{HTML}{d62728}

\usepackage{algpseudocode}
\usepackage{array}
\newcolumntype{L}[1]{>{\raggedright\let\newline\\\arraybackslash\hspace{0pt}}m{#1}}
\newcolumntype{C}[1]{>{\centering\let\newline\\\arraybackslash\hspace{0pt}}m{#1}}
\newcolumntype{R}[1]{>{\raggedleft\let\newline\\\arraybackslash\hspace{0pt}}m{#1}}
\usepackage{eqparbox}
\newdimen{\algindent}
\algnewcommand\LeftComment[2]{%
\hspace{#1\algindent}$\triangleright$ \eqparbox{COMMENT}{#2} \hfill %
}

\usepackage{amsfonts}
\usepackage{xcolor}
\usepackage{mdframed}
\newmdtheoremenv{theo}{Box}
\usepackage{csquotes}
\usepackage{amssymb}

\makeatletter
\renewcommand*{\@seccntformat}[1]{\csname the#1\endcsname.\hspace{0.1cm}}
\makeatother

\title{Meta-Learned Models of Cognition}
\shorttitle{Meta-Learned Models of Cognition}

\authorsnames[1,2,1,2,2,1]{Marcel Binz, Ishita Dasgupta, Akshay Jagadish, Matthew Botvinick, Jane X. Wang, Eric Schulz}
\authorsaffiliations{{Max Planck Institute for Biological Cybernetics},{DeepMind}}

\authornote{
  Correspondence concerning this article should be addressed to Marcel Binz, MPRG Computational Principles of Intelligence, Max Planck  Institute  for  Biological  Cybernetics,  Max Planck Ring 8, 72076 Tübingen, Germany. E-mail: marcel.binz@tue.mpg.de 
    
\addORCIDlink{Marcel Binz}{0000-0001-8872-8386}}
   
\leftheader{Binz}

\abstract{\textbf{Short}: Meta-learning has established itself as a promising tool for building models of human cognition in the recent years. Yet, a coherent research program around meta-learned models of cognition is still missing. The purpose of the present article is to develop such a research program. We accomplish this by pointing out that meta-learning can be used to construct Bayes-optimal learning algorithms, allowing us to draw strong connections to the rational analysis of cognition. We then discuss several advantages of the meta-learning framework over traditional Bayesian methods and reexamine prior work in the context of these new insights. \newline
\textbf{Long}: Meta-learning is a framework for learning learning algorithms through repeated interactions with an environment as opposed to designing them by hand. In recent years, this framework has established itself as a promising tool for building models of human cognition. Yet, a coherent research program around meta-learned models of cognition is still missing. The purpose of this article is to synthesize previous work in this field and establish such a research program. We rely on three key pillars to accomplish this goal. We first point out that meta-learning can be used to construct Bayes-optimal learning algorithms. This result not only implies that any behavioral phenomenon that can be explained by a Bayesian model can also be explained by a meta-learned model but also allows us to draw strong connections to the rational analysis of cognition. We then discuss several advantages of the meta-learning framework over traditional Bayesian methods. In particular, we argue that meta-learning can be applied to situations where Bayesian inference is impossible and that it enables us to make rational models of cognition more realistic, either by incorporating limited computational resources or neuroscientific knowledge. Finally, we reexamine prior studies from psychology and neuroscience that have applied meta-learning and put them into the context of these new insights. In summary, our work highlights that meta-learning considerably extends the scope of rational analysis and thereby of cognitive theories more generally.}


\begin{document}
\maketitle
\setcounter{secnumdepth}{3}

It is hard to imagine cognitive psychology and neuroscience without computational models -- they are invaluable tools to study, analyze, and understand the human mind. Traditionally, such computational models have been hand-designed by expert researchers. In a cognitive architecture, for instance, researchers provide a fixed set of structures and a definition of how these structures interact with each other \citep{anderson2013architecture}. In a Bayesian model of cognition, researchers instead specify a prior and a likelihood function which -- in combination with Bayes’ rule -- fully determine the model's behavior \citep{l2008bayesian}. The framework of meta-learning \citep{bengio1991learning, schmidhuber1987evolutionary, thrun1998learning} offers a radically different approach for constructing computational models by learning them through repeated interactions with an environment instead of requiring an a priori specification from a researcher.

Recently, psychologists have started to apply meta-learning to the study of human learning \citep{griffiths2019doing}. It has been shown that meta-learned models can capture a wide range of empirically observed phenomena that could not be explained otherwise. They, amongst others, reproduce human biases in probabilistic reasoning \citep{dasgupta2020theory}, discover heuristic decision-making strategies used by people \citep{binz2020heuristics}, and generalize compositionally on complex language tasks in a human-like manner \citep{lake2019compositional}. The goal of the present article is to develop a research program around meta-learned models of cognition and, in doing so, offer a synthesis of previous work and outline new research directions. 


To establish such a research program, we will make use of a recent result from the machine learning community showing that meta-learning can be used to construct Bayes-optimal learning algorithms \citep{rabinowitz2019meta, ortega2019meta, mikulik2020meta}. This correspondence is interesting from a psychological perspective because it allows us to connect meta-learning to another already well-established framework: the rational analysis of cognition \citep{anderson2013adaptive, chater1999ten}. In a rational analysis, one first has to specify the goal of an agent along with a description of the environment the agent interacts with. The Bayes-optimal solution for the task at hand is then derived based on these assumptions and tested against empirical data. If needed, assumptions are modified and the whole process is repeated. This approach for constructing cognitive models has had a tremendous impact on psychology because it explains \say{why cognition works, by viewing it as an approximation to ideal statistical inference given the structure of natural tasks and environments} \citep{tenenbaum2021homepage}. The observation that meta-learned models can implement Bayesian inference implies that a meta-learned model can be used as a replacement for the corresponding Bayesian model in a rational analysis and thus suggests that any behavioral phenomenon that can be captured by a Bayesian model can also be captured by a meta-learned model.

\begin{figure}[!ht]
    \centering
    \noindent\includegraphics[]{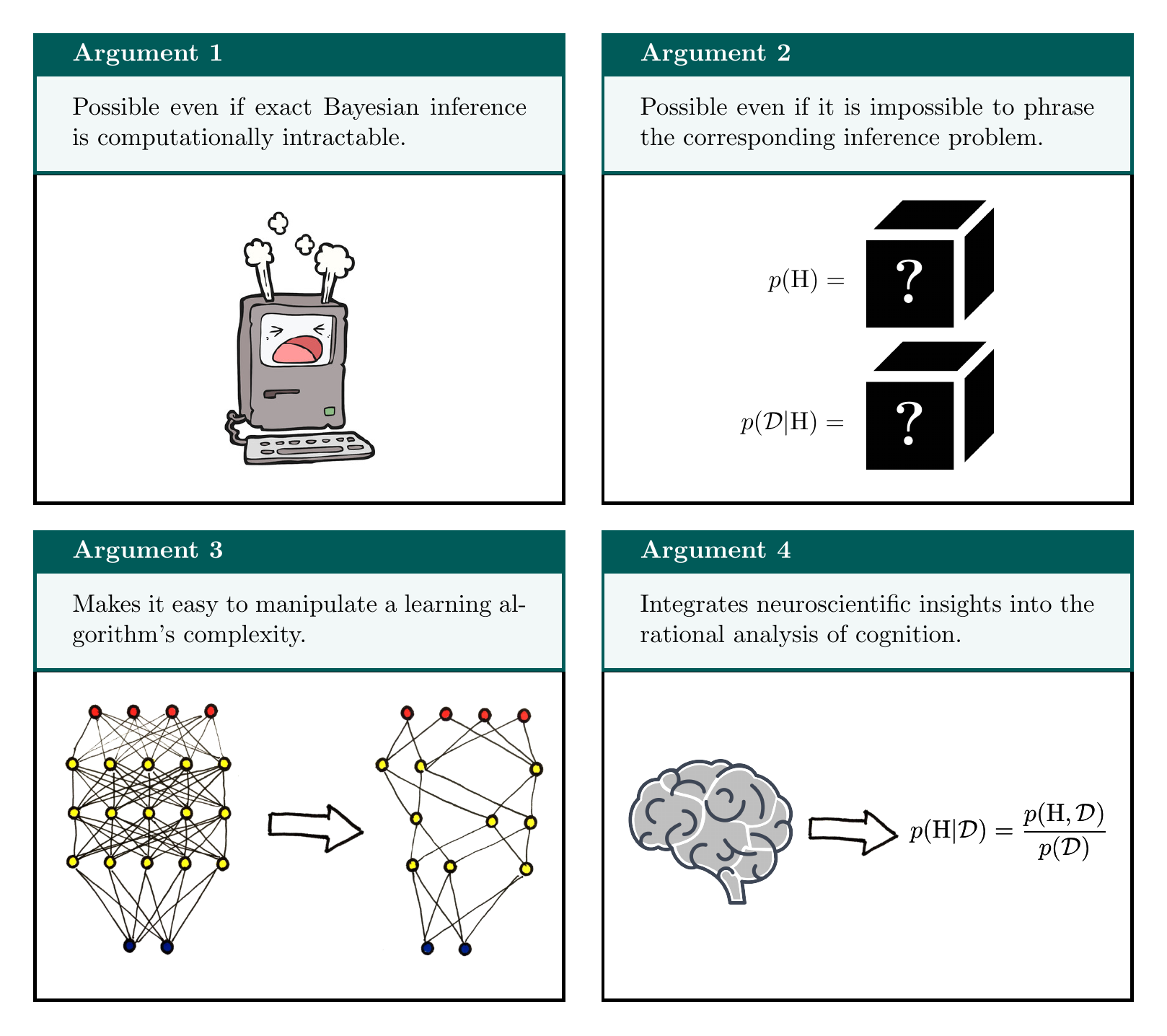}
    \caption{Visual synopsis of the four different arguments for meta-learning over Bayesian inference put forward in this article.}
    \label{fig:summary}
\end{figure}

We start our article by presenting a simplified version of an argument originally formulated by \citet{ortega2019meta} and thereby make their result accessible to a broader audience. Having established that meta-learning produces models that can simulate Bayesian inference, we go on to discuss what additional explanatory power the meta-learning framework offers. After all, why should one not just stick to the tried-and-tested Bayesian approach? We answer this question by providing four original arguments in favor of the meta-learning framework (see Figure \ref{fig:summary} for a visual synopsis):

\begin{enumerate}
\item Meta-learning can produce approximately optimal learning algorithms even if exact Bayesian inference is computationally intractable.
\item Meta-learning can produce approximately optimal learning algorithms even if it is not possible to phrase the corresponding inference problem in the first place. 
\item Meta-learning makes it easy to manipulate a learning algorithm's complexity and can therefore be used to construct resource-rational models of learning.
\item Meta-learning allows us to integrate neuroscientific insights into the rational analysis of cognition by incorporating these insights into model architectures.
\end{enumerate}

The first two points highlight situations in which meta-learned models can be used for rational analysis but traditional Bayesian models cannot. The latter two points provide examples of how meta-learning enables us to make rational models of cognition more realistic, either by incorporating limited computational resources or neuroscientific insights. Taken together, these arguments showcase that meta-learning considerably extends the scope of rational analysis and thereby of cognitive theories more generally. 

We will discuss each of these four points in detail and provide illustrations to highlight their relevance. We then reexamine prior studies from psychology and neuroscience that have applied meta-learning and put them into the context of our newly-acquired insights. For each of the reviewed studies, we highlight how it relates to the four presented arguments, and discuss why its findings could not have been obtained using a classical Bayesian model. Following that, we describe under which conditions traditional models are preferable to those obtained by meta-learning. We finish our article by speculating what the future holds for meta-learning. Therein, we focus on how meta-learning could be the key to building a domain-general model of human cognition.

\section{Meta-Learned Rationality}


The prefix \emph{meta}- is generally used in a self-referential sense: a meta-rule is a rule about rules, a meta-discussion is a discussion about discussions, and so forth. Meta-learning, consequently, refers to learning about learning. We, therefore, need to first establish a common definition of \emph{learning} before covering meta-learning in more detail. For the present article, we adopt the following definition from \citet{mitchell1997machine}: \vspace{0.3cm}


\begin{tcolorbox}[sharp corners, colback=mpigreen!5!white,colframe=mpigreen!75!black,title=\textbf{Definition: Learning}]
  \say{For a given task, training experience, and performance measure, an algorithm is said to learn if its performance at the task improves with experience.}
\end{tcolorbox}


To illustrate this definition, consider the following example which we will return to throughout the text: you are a biologist who has just discovered a new insect species and now set yourself the task of predicting how large members of this species are. You have already observed three exemplars in the wild with lengths of $16$cm, $12$cm, and $15$cm respectively. This data amounts to your training experience. Ideally, you can use this experience to make better predictions about the length of the next individual you encounter. You are said to have learned something if your performance is better after seeing the data than it was before. Typical performance measures for this example problem include the mean squared error or the (negative) log-likelihood. 


\subsection{Bayesian Inference for Rational Analyses}

In a rational analysis of cognition, researchers are trying to compare human behavior to that of an optimal learning algorithm. However, it turns out that no learning algorithm is better than another when averaged over all possible problems \citep{wolpert1996lack, wolpert1997no}, which means that we first have to make additional assumptions about the to-be-solved problem to obtain a well-defined notion of optimality. For our running example, one may make the following -- somewhat unrealistic -- assumptions:

\begin{enumerate}
	\item Each observed insect length $x_k$ is sampled from a normal distribution with mean $\mu$ and standard deviation $\sigma$. 
	\item An insect species' mean length $\mu$ cannot be observed directly, but the standard deviation $\sigma$ is known to be $2$cm.
	\item Mean lengths across all insect species are distributed according to a normal distribution with a mean of $10$cm and a standard deviation of $3$cm. 
\end{enumerate}

An optimal way of making predictions about new observations under such assumptions is specified by Bayesian inference. Bayesian inference requires access to a prior distribution $p(\mu)$ that defines an agent’s initial beliefs about possible parameter values before observing any data and a likelihood $p(x_{1:t} | \mu)$ that captures the agent's knowledge about how data is generated for a given set of parameters. In our running example, the prior and the likelihood can be identified as follows:
\begin{align}
    p(\mu) &= \mathcal{N}(\mu; 10, 3) \label{eq:priord} \\
    p(x_{1:t} | \mu) &= \prod_{k=1}^t p(x_{k} | \mu) = \prod_{k=1}^t \mathcal{N}(x_{k}; \mu, 2) \label{eq:likelihood}
\end{align}
where $x_{1:t} = x_1, x_2, \ldots, x_t$ denotes a sequence of observed insect lengths and the product in Equation \ref{eq:likelihood} arises due to the additional assumption that observations are independent given the parameters.

The outcome of Bayesian inference is a posterior predictive distribution $p(x_{t+1} | x_{1:t})$, which the agent can use to make probabilistic predictions about a hypothetical future observation. To obtain this posterior predictive distribution, the agent first combines prior and likelihood into a posterior distribution over parameters by applying Bayes' theorem:
\begin{equation} \label{eq:pd}
	p(\mu | x_{1:t}) = \dfrac{p(x_{1:t} | \mu)p(\mu)}{\int p(x_{1:t} | \mu)p(\mu) d\mu}
\end{equation} 
In a subsequent step, the agent then averages over all possible parameter values weighted by their posterior probability to get the posterior predictive distribution:
\begin{equation} \label{eq:ppd}
	p(x_{t+1} | x_{1:t}) = \int p(x_{t+1} | \mu) p(\mu | x_{1:t}) d\mu
\end{equation} 



Multiple arguments justify Bayesian inference as a normative procedure, and thereby its use for rational analyses \citep{corner2013normative}. This includes dutch book arguments \citep{lewis_1999, rescorla2020improved}, free energy minimization \citep{hinton1993keeping, friston2010free}, and performance-based justifications \citep{aitchison1975goodness, rosenkrantz1992justification}. For this article, we are mainly interested in the latter class of performance-based justifications because these can be used -- as we will demonstrate later on -- to derive meta-learning algorithms that learn approximations to Bayesian inference. 

Performance-based justifications are based on the notion of frequentist statistics. They assert that no learning algorithm can be better than Bayesian inference on a certain performance measure. Particularly relevant for this article is a theorem first proved by \citet{aitchison1975goodness}. It states that the posterior predictive distribution is the distribution (from the set of all possible distributions $\mathcal{Q}$) that maximizes the log-likelihood of hypothetical future observations when averaged over the data-generating distribution $p(\mu, x_{1:t+1}) = p(\mu) p(x_{1:t+1} | \mu)$:
 \begin{equation} \label{eq:aitchison}
    p(x_{t+1} | x_{1:t}) = \argmax_{q \in \mathcal{Q}} \mathbb{E}_{p(\mu, x_{1:t+1})} \left[ \log q(x_{t+1} | x_{1:t})\right]
\end{equation}
Equation \ref{eq:aitchison} implies that if an agent wants to make a prediction about the length of a still unobserved exemplar from a particular insect species and measures its performance using the log-likelihood, then -- averaged across all possible species that can be encountered -- there is no better way of doing it than using the posterior predictive distribution. We decided to  include a short proof of this theorem within Box $1$ for the curious reader as it does not appear in popular textbooks on probabilistic machine learning \citep{bishop2006pattern, murphy2012machine} nor in survey articles on Bayesian models of cognition. Note that, while the theorem itself is central to our later argument, working through its proof is not required to follow the remainder of this article.




\subsection{Meta-Learning}


\begin{figure}[!ht]
\centering
\hspace*{-0.25cm}\includegraphics[width=1.03\textwidth]{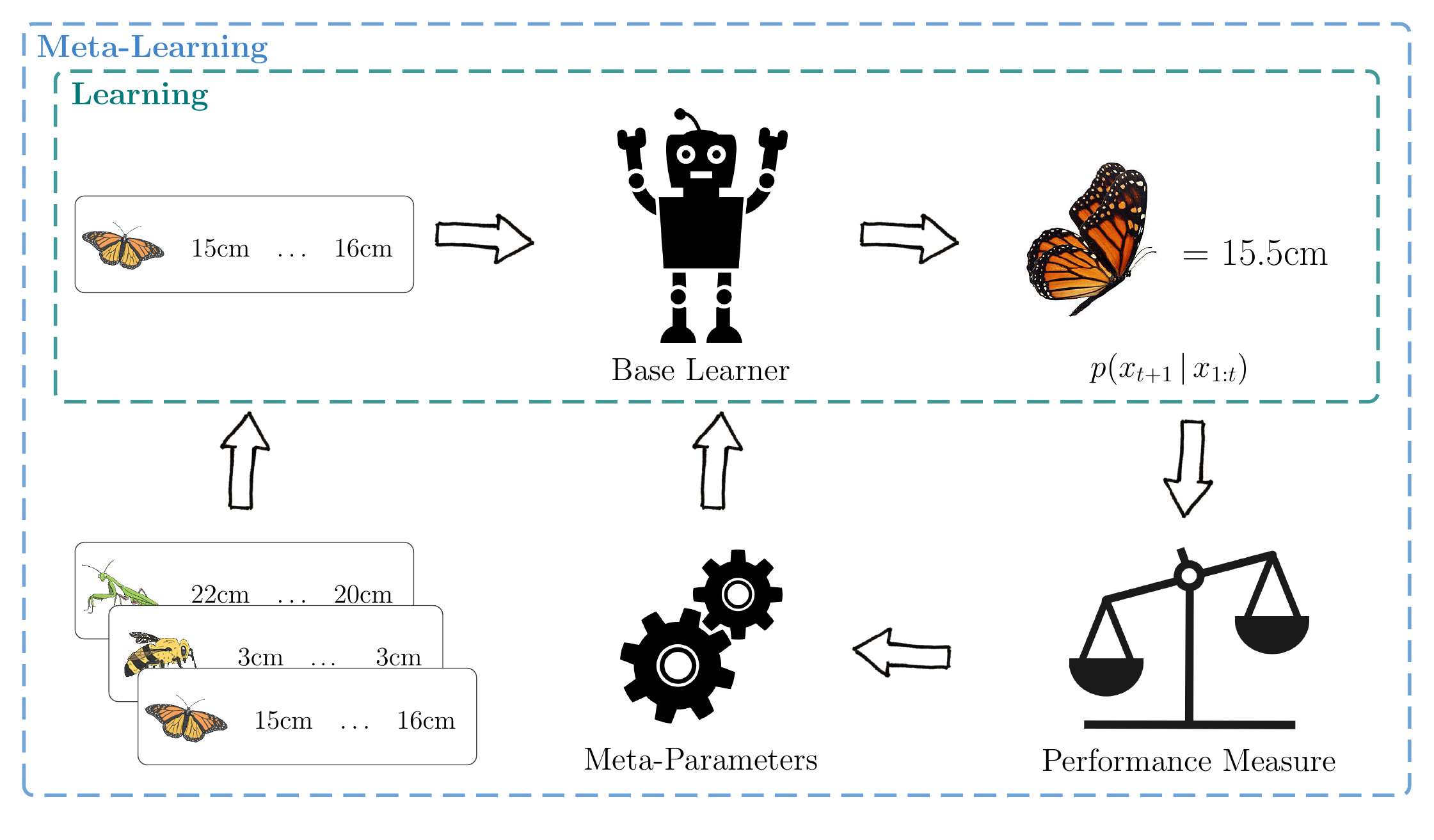}
\caption{High-level overview of the meta-learning process. A base learner (green rectangle) receives data and performs some internal computations that improve its predictions on future data-points. A meta-learner (blue rectangle) encompasses a set of meta-parameters that can be adapted to create an improved learner. This is accomplished by training the learner on a distribution of related learning problems.}
\label{fig:meta}
\end{figure}

Having summarized the general concepts behind Bayes-optimal learning, we can now start to describe meta-learning in more detail. Formally speaking, a meta-learning algorithm is defined as any algorithm that \say{uses its experience to change certain aspects of a learning algorithm, or the learning method itself, such that the modified learner is better than the original learner at learning from additional experience} \citep{Schaul2010Scholarpedia}.

To accomplish this, one first decides on an inner-loop (or base) learning algorithm and determines which of its aspects can be modified. We also refer to these modifiable aspects as meta-parameters. In an outer-loop (or meta-learning) process, the system is then trained on a series of learning problems such that the inner-loop learning algorithm gets better at solving the problems that it encounters. We provide a high-level overview of this framework in Figure~\ref{fig:meta}.

The previous definition is quite broad and includes a variety of methods. It is, for example, possible to meta-learn:
\begin{itemize}
\item Hyperparameters for a base learning algorithm, such as learning rates, batch sizes, or the number of training epochs \citep{doya2002metalearning, li2017meta, feurer2019hyperparameter}.
\item Initial parameters of a neural network that is trained via stochastic gradient descent \citep{finn2017model, nichol2018first}.
\item Prior distributions in a probabilistic graphical model \citep{baxter1998theoretical, grant2018recasting}.
\item Entire learning algorithms \citep{hochreiter2001learning, santoro2016meta}.
\end{itemize}
While all these methods have their own merits, we will be primarily concerned with the latter approach. Learning entire learning algorithms from scratch is arguably the most general and ambitious type of meta-learning, and it is the focus of this article because it is the only one among the aforementioned approaches leading to Bayes-optimal learning algorithms that can be utilized for rational analyses.

\subsection{Meta-Learned Inference}

It may seem like a daunting goal to learn an entire learning algorithm from scratch, but the core idea behind the approach we discuss in the following is surprisingly simple: instead of using Bayesian inference to obtain the posterior predictive distribution, we teach a general-purpose function approximator to do this inference. Previous work has mostly focused on using recurrent neural networks as function approximators in this setting and thus we will -- without loss of generality -- focus our upcoming exposition on this class of models. 

\begin{figure}[!ht]
    \centering
    \hspace*{-1.2cm}\begin{tabular}{@{}cc}
        \textbf{(a) Meta-Learning Setup} & \hspace{-1cm}\textbf{(b) Pseudocode}  \\
        \includegraphics{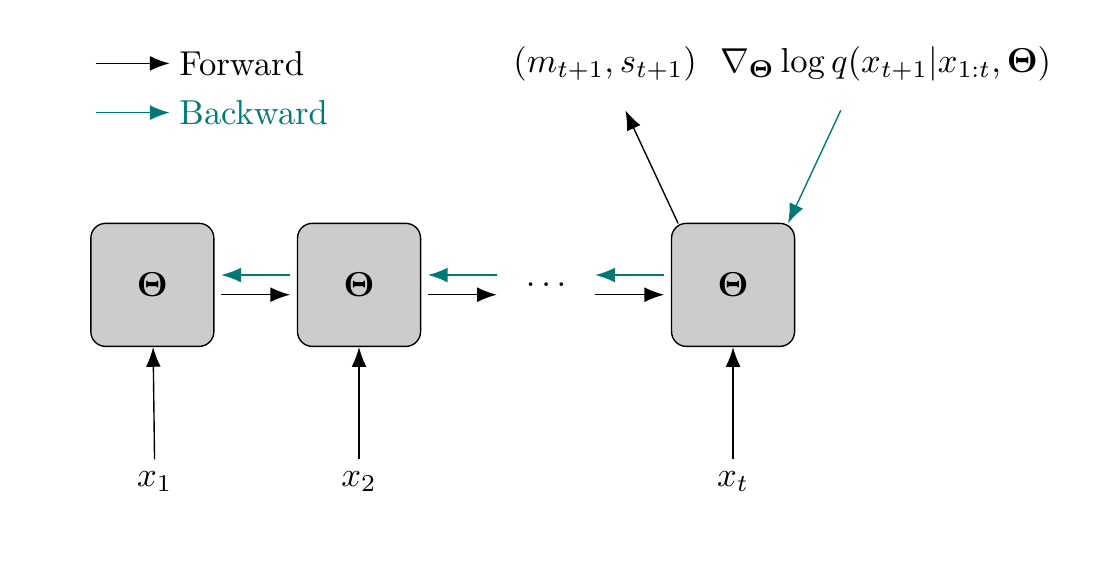} & \raisebox{3.13cm}{{
        \hspace{-1cm}\begin{minipage}{.42\textwidth}
        \begin{algorithmic}
        \State initialize $\mathbf{\Theta}$
        \While{not converged}
        \State \LeftComment{0}{sample data}
        \State $\mu \sim p(\mu)$, $x_{1:t+1} \sim p(x_{1:t+1}| \mu)$
        \State \LeftComment{0}{forward pass}
        \State $q(x_{t+1} | x_{1:t}, \mathbf{\Theta}) \leftarrow$ model($x_{1:t}$)
        \State  \LeftComment{0}{backward pass and update}
        \State $\mathbf{\Theta} \leftarrow \mathbf{\Theta} + \alpha \nabla_{\mathbf{\Theta}}  \log q(x_{t+1} | x_{1:t}, \mathbf{\Theta}) $
        \EndWhile
        \end{algorithmic}
\end{minipage}}} \\
    \end{tabular}
    \hspace*{-0.6cm}\begin{tabular}{@{}cc}
    \textbf{(c) Meta-Learning Loss} &  \textbf{(d) Predictive Distributions} \\
    \includegraphics{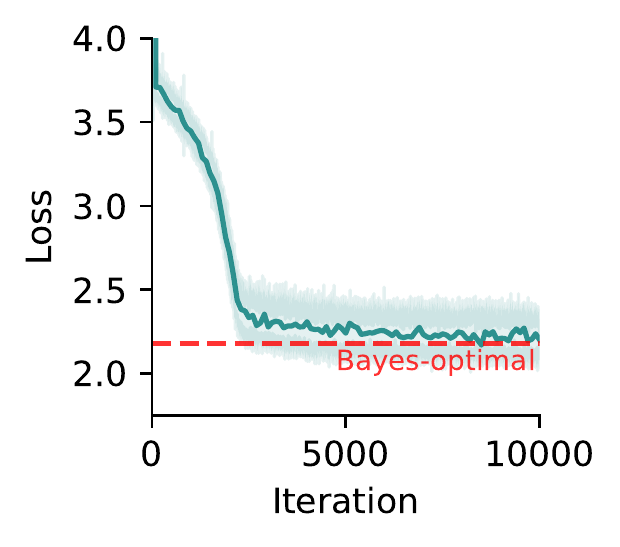} & \includegraphics[]{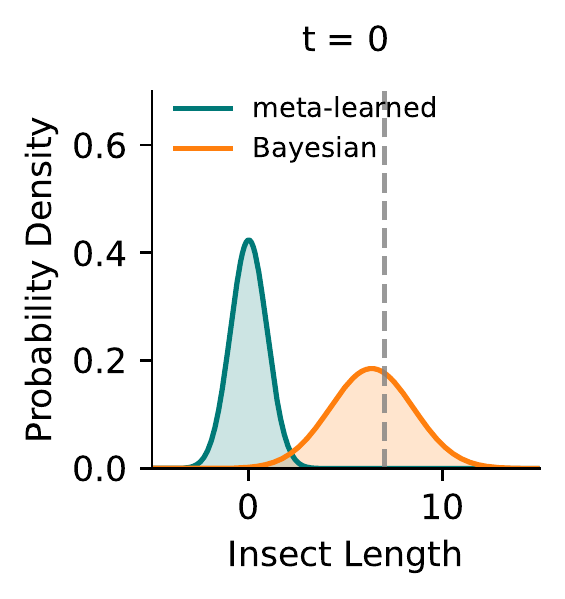} \hspace*{-0.cm}\includegraphics[]{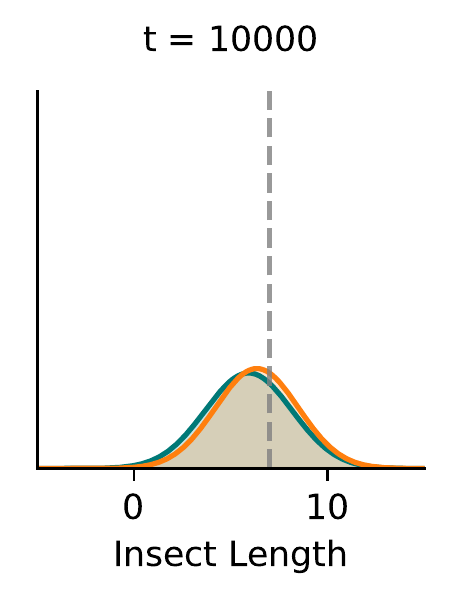} 
    \end{tabular}
    \caption{Meta-learning illustration. (a) A recurrent neural network processes a sequence of observations and produces a predictive distribution at the final time-step. (b) Pseudocode for a simple meta-learning algorithm. (c) Loss during meta-learning with shaded contours corresponding to the standard deviation across 30 runs. (d) Posterior and meta-learned predictive distributions for an example sequence at beginning and end of meta-learning. The dotted grey line denotes the (unobserved) mean length.}
    \label{fig:performance}
\end{figure}

Like the posterior predictive distribution, the recurrent neural network processes a sequence of observed length from a particular insect species and produces a predictive distribution over the lengths of potential future observations from the same species. More concretely, the meta-learned predictive distribution takes a predetermined functional form whose parameters are given by the network outputs. If we had, for example, decided to use a normal distribution as the functional form of the meta-learned predictive distribution, outputs of the network would correspond to a expected length $m_{t+1}$ and its standard deviation $s_{t+1}$. Figure \ref{fig:performance}a illustrates this setup graphically.

Initially, the recurrent neural network implements a randomly initialized learning algorithm.\footnote{Based on our earlier definition, it is at this point strictly speaking not a learning algorithm at all as it does not improve with additional data.} The goal of the meta-learning process is then to turn this system into an improved learning algorithm. The final result is a learning algorithm that is \emph{learned} or trained rather than specified by a practitioner. To create a learning signal to do this training, we need a performance measure that can be used to optimize the network. Equation \ref{eq:aitchison} suggests a straightforward strategy for designing such a measure by replacing the maximization over all possible distributions with a maximization over meta-parameters $\mathbf{\Theta}$ (in our case, the weights of the recurrent neural network):
\begin{align}
    &\argmax_{q \in \mathcal{Q}} \mathbb{E}_{p(\mu, x_{1:t+1})} \left[ \log q(x_{t+1} | x_{1:t})\right] \nonumber \\
    \approx ~  &\argmax_{\mathbf{\Theta}}\mathbb{E}_{p(\mu, x_{1:t+1})} \left[ \log q(x_{t+1} | x_{1:t}, \mathbf{\Theta})\right] \label{eq:ppdmeta}
\end{align}
To turn this expression into a practical meta-learning algorithm, we will -- as common practice when training deep neural networks -- maximize a sample-based version using stochastic gradient ascent:
\begin{align}
    &\argmax_{\mathbf{\Theta}}\mathbb{E}_{p(\mu, x_{1:t+1})} \left[ \log q(x_{t+1} | x_{1:t}, \mathbf{\Theta})\right] \nonumber \\
    \approx ~  &\argmax_{\mathbf{\Theta}} \frac{1}{N}\sum_{n=1}^{N} \log q(x_{t+1}^{(n)} | x_{1:t}^{(n)}, \mathbf{\Theta})  \label{eq:ppdmetasample} 
\end{align}

Figure \ref{fig:performance}b presents pseudocode for a simple gradient-based procedure that maximizes Equation \ref{eq:ppdmetasample}. The entire meta-learning algorithm can be implemented in just around $30$ lines of self-contained \texttt{PyTorch} code \citep{NEURIPS2019_9015}. We provide an annotated reference implementation on this article's accompanying github repository.\footnote{\url{https://github.com/marcelbinz/meta-learned-models}}





\subsection{How Good Is a Meta-Learned Algorithm?}

We have previously shown that the global optimum of Equation \ref{eq:ppdmetasample} is achieved by the posterior predictive distribution. Thus, by maximizing this performance measure, the network is actively encouraged to implement an approximation to exact Bayesian inference. Importantly, after the completion of meta-learning, producing an approximation to the posterior predictive distribution does not require any further updates to the network weights. To perform an inference, we simply have to query the network's outputs after providing it with a particular sequence of observations. 

If we want to use the fully optimized network for rational analyses, we have to ask ourselves: how well does the resulting model approximate Bayesian inference? Two aspects have to be considered when answering this question. First, the network has to be sufficiently expressive to produce the exact posterior predictive distribution for all input sequences. Neural networks of sufficient width are universal function approximators \citep{hornik1989multilayer}, meaning that they can approximate any continuous function to arbitrary precision. Therefore, this aspect is not too problematic for the optimality argument. The second aspect is a bit more intricate: assuming that the network is powerful enough to represent the global optimum of Equation \ref{eq:ppdmetasample}, the employed optimization procedure also has to find it. While we are not aware of any theorem that could provide such a guarantee, in practice, it has been observed that meta-learning procedures similar to the one discussed here often lead to networks that closely approximate Bayesian inference \citep{NEURIPS2020_d902c3ce, rabinowitz2019meta}. We provide a visualization demonstrating that the predictions of a meta-learned model closely resemble those of exact Bayesian inference for our insect length example in Figure \ref{fig:performance}c-d. 

While our exposition in this section focused on the supervised learning case, the same ideas can also be readily extended to the reinforcement learning setting \citep{wang2016learning, duan2016rl}. Box $2$ outlines the general ideas behind the meta-reinforcement learning framework.

\subsection{Tool or Theory?}

It is often not so trivial to separate meta-learning from normal learning. We believe that part of this confusion arises from an underspecification regarding what is being studied. In particular, the meta-learning framework provides opportunities to address two distinct research questions:
\begin{enumerate}
    \item It can be used to study how people improve their learning abilities over time.
    \item It can be used as a methodological tool to construct learning algorithms with the properties of interest (and thereafter compare the emerging learning algorithms to human behavior).
\end{enumerate}
Historically, behavioral psychologists have been mainly interested in the former aspect \citep{harlow1949formation, doya2002metalearning}. In the 1940s, for example, \citet{harlow1949formation} already studied how learning in monkeys improves over time. He found that they adapted their learning strategies after sufficiently many interactions with tasks that shared a common structure, thereby showing a learning-to-learn effect. By now, examples of this phenomenon have been found in many different species -- including humans -- across nature \citep{wang2021meta}.

More recently, psychologists have started to view meta-learning as a methodological tool to construct approximations to Bayes-optimal learning algorithms \citep{binz2020heuristics, kumar2020meta}, and subsequently utilize the resulting algorithms to study human cognition. The key difference from the former approach is that, in this setting, one abstracts away from the process of meta-learning and instead focuses on its outcome. From this perspective, only the fully converged model is of interest. Importantly, this approach allows us to investigate human learning from a rational perspective since we have demonstrated that meta-learning can be used to construct approximations to Bayes-optimal learning.

We place an emphasis on the second aspect in the present article and advocate for using fully converged meta-learned algorithms -- as replacements for the corresponding Bayesian models -- for rational analyses of cognition.\footnote{There has been a longstanding conceptual debate in cognitive psychology on whether to view Bayesian models as normative standards or descriptive tools. We believe that this debate is beyond the scope of the current article and thus refer the reader to earlier work for an in-depth discussion \citep{jones2011bayesian, griffiths2012bayesians, zednik2016bayesian, tauber2017bayesian}. We only want to add that the framework outlined here is agnostic to this issue -- meta-learned models may serve as both normative standards and descriptive tools.} In the next section, we will outline several arguments that support this approach. However, it is important to mention that we believe that meta-learning can also be a valuable tool to understand the process of learning-to-learn itself. In this context, several intriguing questions arise: at what time scale does meta-learning take place in humans? How much of it is due to task-specific adaptations? How much of it is based on evolutionary or developmental processes? While we agree that these are important questions, they are not the focus of this article.

\section{Why Not Bayesian Inference?} \label{sec:why}

We have just argued that it is possible to meta-learn Bayes-optimal learning algorithms. What are the implications of this result? If one has access to two different theories that make identical predictions, which of them should be preferred? Bayesian inference has already established itself as a valuable tool for building cognitive models in the recent decades. Thus, the burden of proof is arguably on the meta-learning framework. In this section, we provide four different arguments that highlight the advantages of meta-learning for building models of cognition. Many of these arguments are novel and have not been put forward explicitly in previous literature. The first two arguments highlight situations in which meta-learned models can be used for rational analysis but traditional Bayesian models cannot. The latter two provide examples of how meta-learning enables us to make rational models of cognition more realistic, either by incorporating limited computational resources or neuroscientific insights.

\subsection{Intractable Inference} 

\begin{tcolorbox}[sharp corners, colback=mpigreen!5!white,colframe=mpigreen!75!black,title=\textbf{Argument 1}]
  Meta-learning can produce approximately optimal learning algorithms even if exact Bayesian inference is computationally intractable.
\end{tcolorbox}

Bayesian inference becomes intractable very quickly because the complexity of computing the normalization constant that appears in the denominator grows exponentially with the number of unobserved parameters. In addition, it is only possible to find a closed-form expression of the posterior distribution for certain combinations of prior and likelihood. In our running example, we assumed that both prior and likelihood follow a normal distribution, which, in turn, leads to a normally-distributed posterior. However, if one would instead assume that the prior over mean length follows an exponential distribution -- which is arguably a more sensible assumption as it enforces lengths to be positive -- it becomes already impossible to find an analytical expression for the posterior distribution.

Researchers across disciplines have recognized these challenges and have, in turn, developed approaches that can approximate Bayesian inference without running into computational difficulties. Prime examples of this are variational inference \citep{jordan1999introduction} and Markov chain Monte-Carlo (MCMC) methods \citep{geman1984stochastic}. In variational inference, one phrases inference as an optimization problem by positing a variational approximation whose parameters are fitted to minimize a divergence measure to the true posterior distribution. MCMC methods, on the other hand, draw samples from a Markov chain that has the posterior distribution as its equilibrium distribution. Previous research in cognitive science indicates that human learning shows characteristics of such approximations \citep{daw2008semi, courville2008rat, sanborn2010rational, sanborn2013constraining, dasgupta2017hypotheses}. 

Meta-learned inference also never requires an explicit calculation of the exact posterior or posterior predictive distribution. Instead, it performs approximately optimal inference via a single forward pass through the network. Inference, in this case, is approximate because we had to determine a functional form for the predictive distribution. The chosen form may deviate from the true form of the posterior predictive distribution, which, in turn, leads to approximation errors.\footnote{In principle, one could select arbitrarily flexible functional forms, such as mixtures of normal distributions or discretized distributions with small bin sizes, which would reduce the accompanying approximation error.} In some sense, this type of approximation is similar to variational inference: both approaches involve optimization and require one to define a functional form of the respective distribution. However, the optimization process in both approaches uses a different loss function and happens at different time scales. While variational inference employs the negative evidence lower bound as its loss function, meta-learning directly maximizes for models which can be expected to generalize well to unseen observations (using the performance-based measure from Equation \ref{eq:aitchison}). Furthermore, meta-learned inference only involves optimization during the outer-loop meta-learning process but not during the actual learning itself. To update how a meta-learned model makes predictions in the light of new data, we only have to perform a simple forward pass through the network. In contrast to this, standard variational inference requires us to rerun the whole optimization process from scratch every time a new data point is observed.\footnote{This only holds for standard variational inference but not for more advanced methods that involve amortization such as variational autoencoders \citep{kingma2013auto}.}

In summary, it is possible to meta-learn an approximately Bayes-optimal learning algorithm. If exact Bayesian inference is not tractable, such models are our best option for performing rational analyses. Yet, many other methods for approximate inference, such as variational inference and MCMC methods, also share this feature, and it will thus ultimately be an empirical question which of these approximations provides a better description of human learning.


\subsection{Unspecified Problems}

\begin{tcolorbox}[sharp corners, colback=mpigreen!5!white,colframe=mpigreen!75!black,title=\textbf{Argument 2}]
  Meta-learning can produce optimal learning algorithms even if it is not possible to phrase the corresponding inference problem in the first place.
\end{tcolorbox}

Bayesian inference is hard, but posing the correct inference problem can be even harder. What exactly do we mean by that? To perform Bayesian inference, we need to specify a prior and a likelihood. Together, these two objects fully specify the assumed data-generating distribution, and thus the inference problem. Ideally, the specified data-generating distribution should match how the environment actually generates its data. It is fairly straightforward to fulfill this requirement in artificial scenarios, but for many real-world problems, it is not. Take for instance our running example: Does the prior over mean length really follow a normal distribution? If yes, what are the mean and variance of this distribution? Are the underlying parameters actually time-invariant or do they, for example, change based on seasons? None of these questions can be answered with certainty.

In his seminal work on Bayesian decision theory, \citet{savage1972foundations} made the distinction between small and large world problems. A small world problem is one \say{in which all relevant alternatives, their consequences, and probabilities are known} \citep{gigerenzer2011heuristic}. A large world problem, on the other hand, is one in which the prior, the likelihood, or both cannot be identified. Savage's distinction between small and large worlds is relevant for the rational analysis of human cognition as its critics have pointed out that Bayesian inference only provides a justification for optimal reasoning in small world problems \citep{binmore2007rational} and that \say{very few problems of interest to the cognitive, behavioral, and social sciences can be said to satisfy [this] condition} \citep{brighton2012rational}.

Identifying the correct set of assumptions becomes especially challenging once we deal with more complex problems. To illustrate this, consider a study conducted by \citet{lucas2015rational} who attempted to construct a Bayesian model of human function learning. Doing so required them to specify a prior over functions that people expect to encounter. Without direct access to such a distribution, they instead opted for a heuristic solution: $98.8$\% of functions are expected to be linear, $1.1$\% are expected to be quadratic, and $0.1$\% are expected to be non-linear. Empirically, this choice led to good results, but it is hard to justify from a rational perspective. We simply do not know the frequency with which these functions appear in the real world, nor whether the given selection fully covers the set of functions expected by participants.


There are also inference problems in which it is not possible to specify or compute the likelihood function. These problems have been studied extensively in the machine learning community under the names of simulation-based or likelihood-free inference \citep{cranmer2020frontier, lueckmann2021benchmarking}. In this setting, it is typically assumed that we can sample data from the likelihood for a given parameter setting but that computing the corresponding likelihood is impossible. Take, for instance, our insect length example. It should be clear that an insect’s length does not only depend on its species’ mean but on many other factors such as climate, genetics, and the individual's age. Even if all these factors were known, mapping them to a likelihood function does seem close to impossible. But, we can generate samples easily by observing insects in the wild. If we had access large database of insect length measurements for different species, this could be directly used to meta-learn an approximately Bayes-optimal learning algorithm for predicting their length, while circumventing an explicit definition of a likelihood function.

In cases where we do not have access to a prior or a likelihood, we can neither apply exact Bayesian inference nor approximate inference schemes such as variational inference or MCMC methods. In contrast to this, meta-learned inference does not require us to define the prior or the likelihood explicitly. It only demands samples from the data-generating distribution to meta-learn an approximately Bayes-optimal learning algorithm -- a much weaker requirement \citep{muller2021transformers}. The ability to construct Bayes-optimal learning algorithms for large worlds problems is a unique feature of the meta-learning framework, and we believe that it could open up totally new avenues for constructing rational models of human cognition. To highlight one concrete example, it would be possible to take a collection of real-world decision-making tasks -- such as the ones presented by \citet{czerlinski1999good} -- and use them to obtain a meta-learned agent that is adapted to the decision-making problems that people actually encounter in their everyday lives. This algorithm could then serve as a normative standard against which we can compare human decision-making.


\subsection{Resource Rationality} \label{sec:rr}

\begin{tcolorbox}[sharp corners, colback=mpigreen!5!white,colframe=mpigreen!75!black,title=\textbf{Argument 3}]
  Meta-learning makes it easy to manipulate a learning algorithm's complexity and can therefore be used to construct resource-rational models of learning.
\end{tcolorbox}

Bayesian inference has been successfully applied to model human behavior across a number of domains, including perception \citep{knill1996perception}, motor control \citep{kording2004bayesian}, everyday judgments \citep{griffiths2006optimal}, and logical reasoning \citep{oaksford2007bayesian}. Notwithstanding these success stories, there are also well-documented deviations from the notion of optimality prescribed by Bayesian inference. People, for example, underreact to prior information \citep{kahneman1973psychology}, ignore evidence \citep{benjamin2019errors}, and rely on heuristic decision-making strategies \citep{gigerenzer2011heuristic}. 


The intractability of Bayesian inference -- together with empirically observed deviations from it -- has led researchers to conjecture that people only attempt to approximate Bayesian inference. Many different notions of what constitutes a computational resource have been suggested, such as memory \citep{dasgupta2021memory}, thinking time \citep{ratcliff2008diffusion}, or physical effort \citep{hoppe2016learning}. 

\citet{cover1999elements} put forward a dichotomy that will be useful for our following discussion. He refers to the algorithmic complexity of an algorithm as the number of bits needed to \emph{implement} it. In contrast, he refers to the computational complexity of an algorithm as the space, time, or effort required to \emph{execute} it. It is possible to cast many approximate inference schemes as resource-rational algorithms \citep{sanborn2017types}. The resulting models typically consider some form of computational complexity. In MCMC methods, computational complexity can be measured in terms of the number of drawn samples: drawing fewer samples leads to faster inference at the cost of introducing a bias \citep{courville2008rat, sanborn2010rational}. In variational inference, on the other hand, it is possible to introduce an additional parameter that allows to trade-off performance against the computational complexity of transforming the prior into the posterior distribution \citep{ortega2015information, binz2021reconstructing}. Likewise, other frameworks for building resource-rational models, such as rational meta-reasoning \citep{lieder2017strategy}, also only target computational complexity.

The prevalence of resource-rational models based on computational complexity is likely due to the fact that building similar models based on algorithmic complexity is much harder. Measuring algorithmic complexity historically relies on the notion of Kolmogorov complexity, which is the size of the shortest computer program that produces a particular data sequence \citep{solomonoff1964formal, kolmogorov1965three, chaitin1969simplicity}. Kolmogorov complexity is in general uncomputable, and, therefore, of limited practical interest.\footnote{Having said that, it is possible to approximate it under certain circumstances and different authors have applied such approximations to study both human and animal cognition \citep{chater2003simplicity, gauvrit2014algorithmic, zenil2015approximations, gauvrit2017human, griffiths2018subjective}.}


Meta-learning provides us with a straightforward way to manipulate both algorithmic and computational complexity in a common framework by adapting the size of the underlying neural network model. Limiting the complexity of network weights places a constraint on algorithmic complexity (as reducing the number of weights decreases the amount of bits needed to store them, and hence also the amount of bits needed to store the learning algorithm). Limiting the complexity of activations, on the other hand, places a constraint on computational complexity (reducing the number of hidden units, for example, decreases the memory needed for executing the meta-learned model).


Previously, both forms of complexity constraints have been realized in meta-learned models. \citet{dasgupta2020theory} decreased the number of hidden units of a meta-learned inference algorithm, effectively reducing its computational complexity. In contrast, \citet{binz2020heuristics} placed a constraint on the description length of neural network weights, which implements a form of algorithmic complexity. To the best of our knowledge, no other class of resource-rational models exists that allows us to take both algorithmic and computational complexity into account, making this ability a unique feature of the meta-learning framework.

\subsection{Neuroscience} \label{sec:neuro}

\begin{tcolorbox}[sharp corners, colback=mpigreen!5!white,colframe=mpigreen!75!black,title=\textbf{Argument 4}]
  Meta-learning allows us to integrate neuroscientific insights into the rational analysis of cognition by incorporating these insights into model architectures.
\end{tcolorbox}



In addition to providing a framework for understanding many aspects of behavior, meta-learning offers a powerful lens through which to view brain structure and function. For instance, \citet{wang2018prefrontal} presented observations supporting the hypothesis that prefrontal circuits may constitute a meta-reinforcement learning system. From a computational perspective, meta-learning strives to learn a faster inner-loop learning algorithm via an adjustment of neural network weights in a slower outer-loop learning process. Within the brain, an analogous process plausibly occurs when slow, dopamine-driven synaptic change gives rise to reinforcement learning processes that occur within the activity dynamics of the prefrontal network, allowing for adaptation on much faster timescales. This perspective recontextualized the role of dopamine function in reward-based learning and was able to account for a range of previously puzzling neuroscientific findings. To highlight one example, \citet{bromberg2010pallidus} found that dopamine signaling reflected updates in not only \textit{experienced} but also \textit{inferred} values of targets. Notably, a meta-reinforcement learning agent trained on the same task also recovered this pattern. Having a mapping of meta-reinforcement learning components onto existing brain regions furthermore allows us to apply experimental manipulations that directly perturb neural activity, for example by using optogenetic techniques. \citet{wang2018prefrontal} used this idea to modify their original meta-reinforcement learning architecture to mimic the blocking or enhancement of dopaminergic reward prediction error signals, in direct analogy with optogenetic stimulation delivered to rats performing a two-armed bandit task \citep{stopper2014overriding}. 



Importantly, the direction of exchange can also work in the other direction, with neuroscientific findings constraining and inspiring new forms of meta-learning architectures. \citet{bellec2018long}, for example, showed that recurrent networks of spiking neurons are able to display convincing learning-to-learn behavior, including in the realm of reinforcement learning. Episodic meta-reinforcement learning \citep{ritter2018been} architectures are also heavily inspired by neuroscientific accounts of complementary learning systems in the brain \citep{mcclelland1995there}. Both of these examples demonstrate that meta-learning can be used to build more biologically plausible learning algorithms, and thereby highlight that it can act as a bridge between Marr's computational and implementational level \citep{marr2010vision}.


Finally, the meta-learning perspective not only allows us to connect machine learning and neuroscience via architectural design choices but also via the kinds of tasks that are of interest. \citet{dobs2022brain}, for instance, suggested that functional specialization in neural circuits, which has been widely observed in biological brains, arises spontaneously as a consequence of task demands. In particular, they found that convolutional neural networks trained on both face and object recognition depicted emergent segregation on the basis of these tasks. Likewise, \citet{yang2019task} found that training a single recurrent neural network to perform a wide range of cognitive tasks yielded units that were clustered along different functional cognitive processes. Put another way, it seems plausible that functional specialization emerges by training neural networks on multiple tasks. While this has not been tested so far, we speculate that this also holds in the meta-learning setting, as it involves training on multiple tasks by design. If this were true, we could look at the emerging areas inside a meta-learned model, and use the resulting insights to generate novel predictions about the processes happening in individual brain areas \citep{kanwisher2023using}.



\section{Previous Research}

Meta-learned models are already starting to transform the cognitive sciences today. They allow us to model things that are hard to capture with traditional models such as compositional generalization, language understanding, and model-based reasoning. In this section, we provide an overview of what has been achieved with the help of meta-learning in previous work. We arranged this review into various thematic subcategories. For each of them, we summarize which key findings have been obtained by meta-learning and discuss why these results would have been difficult to obtain using traditional models of learning by appealing to the insights from the previous section.

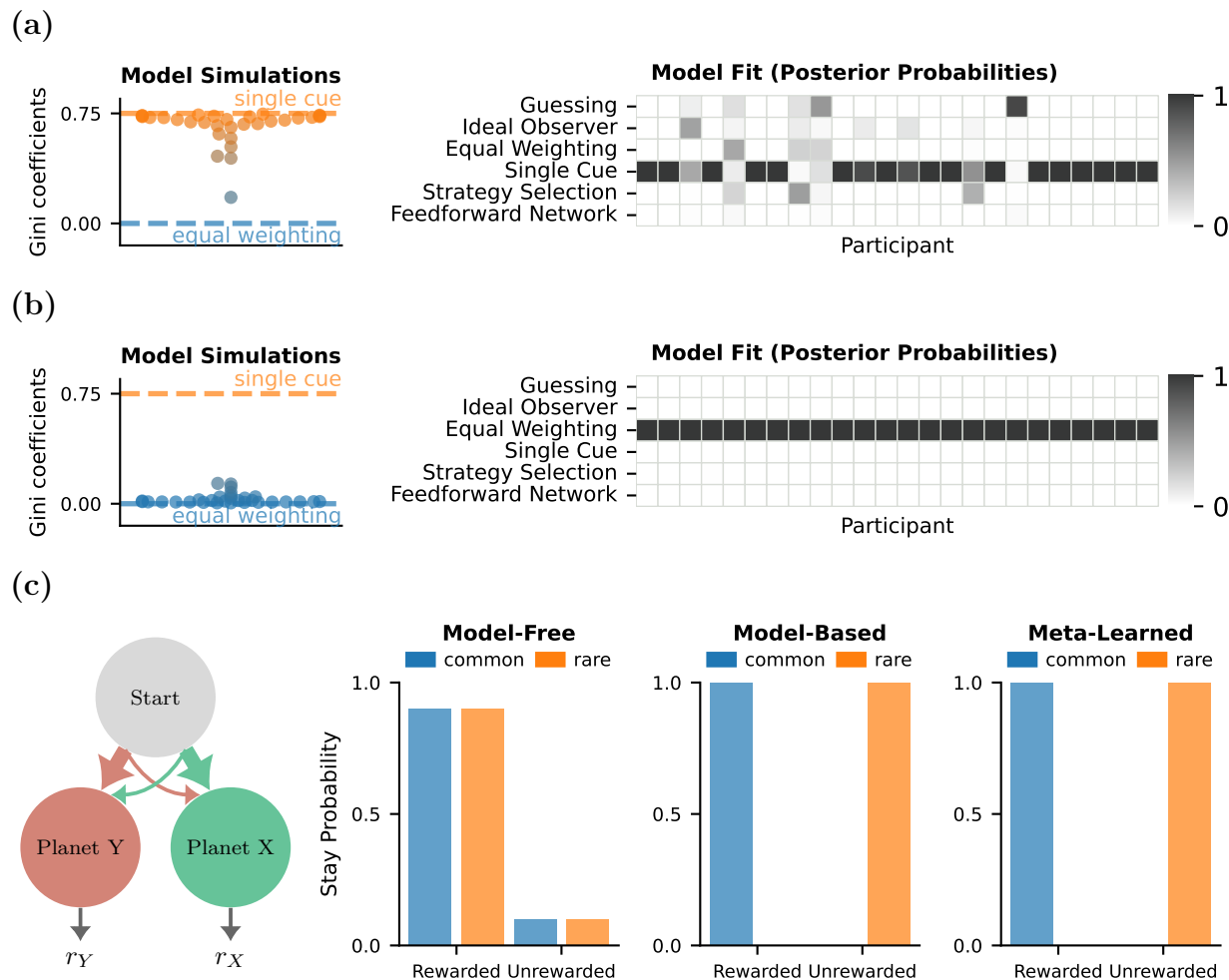
\begin{figure}[!h]
    \flushleft
    \textbf{(a)} \\
    \includegraphics{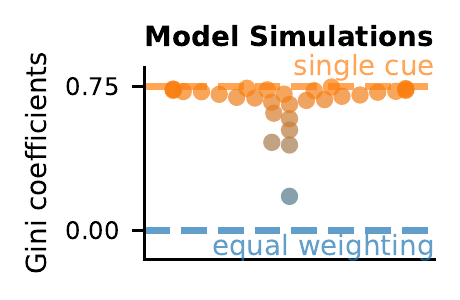} \includegraphics{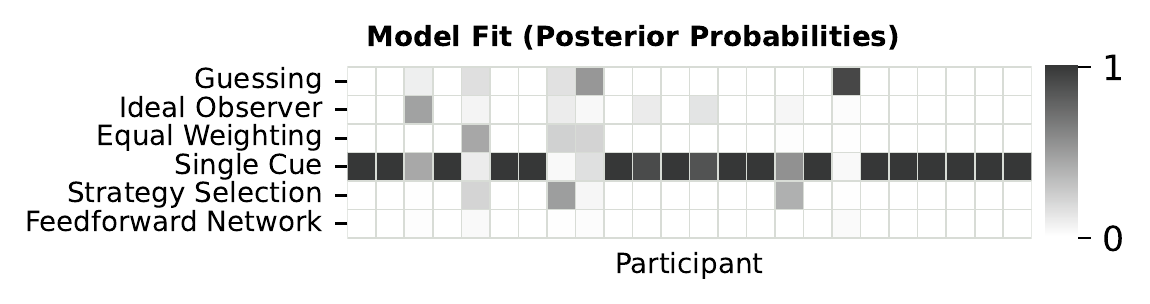}\\
    \textbf{(b)} \\
    \includegraphics{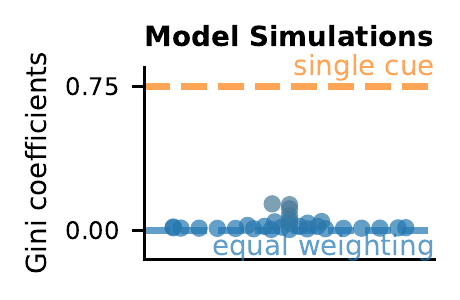}
    \includegraphics{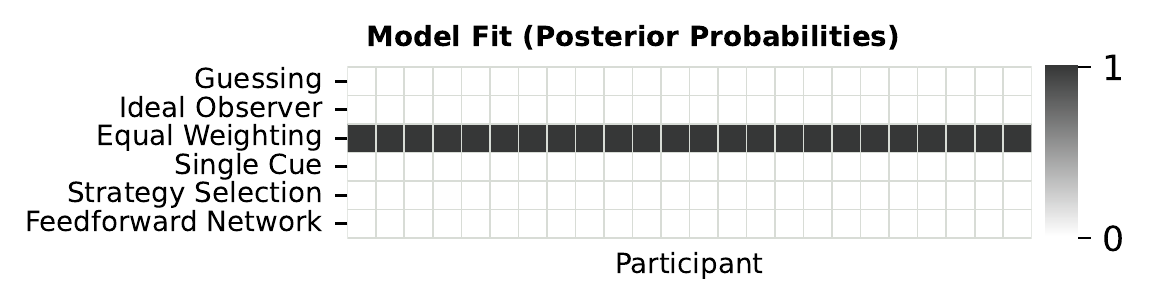}\\
    \textbf{(c)} \\ 
    
    \centering
    \raisebox{0.3cm}{\begin{tikzpicture}
        \node[circle, minimum width=1.6cm, fill=gray!30] at (0, 0) (a1) {\scriptsize Start};
        \node[circle, minimum width=1.6cm, fill=BrickRed!60] at (-1., -2) (a2) {\scriptsize Planet Y};
        \node[, circle, minimum width=1.6cm, fill=ForestGreen!60] at (1., -2) (a3) {\scriptsize Planet X};
        \node[] at (-1., -3.5) (a4) {\footnotesize $r_Y$};
        \node[] at (1., -3.5) (a7) {\footnotesize $r_X$};

        \draw [-{Latex[length=4mm, width=6mm]}, line width=2mm, BrickRed!60] (a1.240) to (a2.70);
        \draw [-{Latex[length=4mm, width=6mm]}, line width=2mm, ForestGreen!60] (a1.300) to (a3.110);
         \draw [-{Latex[length=2mm, width=3mm]}, line width=0.5mm, BrickRed!60] (a1.240) to[bend right]  (a3.120);
        \draw [-{Latex[length=2mm, width=3mm]}, line width=0.5mm, ForestGreen!60] (a1.300) to[bend left]  (a2.60);
        
        \draw [-{Latex[length=2mm, width=3mm]}, line width=0.5mm, black!60] (a2) to  (a4);
        \draw [-{Latex[length=2mm, width=3mm]}, line width=0.5mm, black!60] (a3) to  (a7);
        
        \end{tikzpicture}}
    \includegraphics{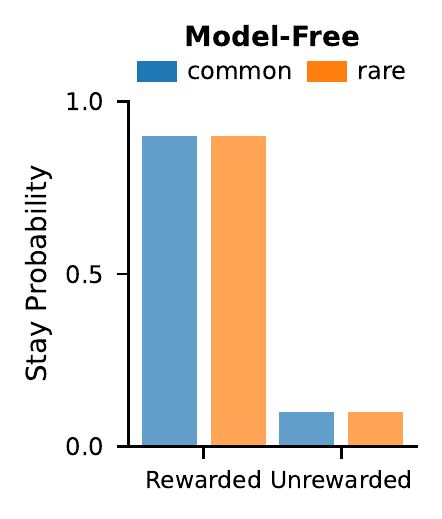} \hspace*{-0.2cm}\includegraphics{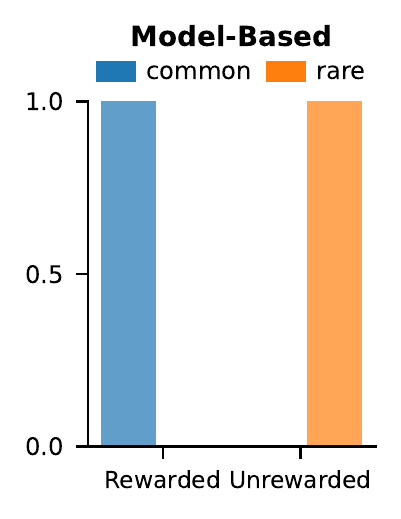} \hspace*{-0.2cm}\includegraphics{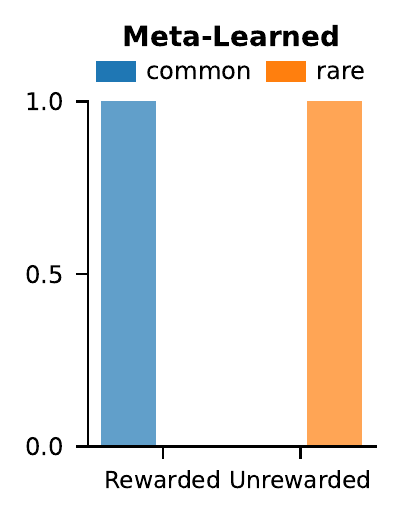}
    \caption{Example results obtained using meta-learned models. (a) In a paired comparison task, a meta-learned model identified a single-cue heuristic as the resource-rational solution when information about the feature ranking was available. Follow-up experiments revealed that people indeed apply this heuristic under the given circumstances. (b) If information about feature directions was available, the same meta-learned model identified an equal weighting heuristic as the resource-rational solution. People also applied this heuristic in the given context \citep{binz2020heuristics}. (c) \citet{wang2016learning} showed that meta-learned models can exhibit model-based learning characteristics in the two-step task \citep{daw2011model} even when they were purely trained through model-free approaches. The plots on the right illustrate the probability of repeating the previous action for different agents (model-free, model-based, meta-learned) after a common or uncommon transition and after a received or omitted reward.}
    \label{fig:examples}
\end{figure}

\subsection{Heuristics and Cognitive Biases}


Meta-learning has been previously used to discover algorithms with a limited computational budget that show human-like cognitive biases as we have already alluded to earlier. \citet{dasgupta2020theory} trained a neural network on a distribution of probabilistic inference problems while controlling for the number of its hidden units. They found that their model -- when restricted to just a single hidden unit -- captured many biases in human reasoning, including a conservatism bias and base rate neglect. Likewise, \citet{binz2020heuristics} trained a neural network on a distribution of decision-making problems while controlling for the number of bits needed to represent the network. Their model discovered two previously suggested heuristics in specific environments and made precise prognoses about when these heuristics should be applied. In particular, knowing the correct ranking of features led to one reason decision-making, knowing the directions of features led to an equal weighting heuristic, and not knowing about either of them led to strategies that use weighted combinations of features (also see Figure~\ref{fig:examples}a-b). 

In both of these studies, meta-learned models offered a novel perspective on results that were previously viewed as contradictory. This was in part possible because meta-learning enabled us to easily manipulate the complexity of the underlying learning algorithm. While doing so is, at least in theory, also possible within the Bayesian framework, no Bayesian model that captures the full set of findings from \citet{dasgupta2020theory} and \citet{binz2020heuristics} has been discovered so far. We hypothesize that this could be because traditional rational process models struggle to capture that human strategy selection is context-dependent even before receiving any direct feedback signal \citep{mercier2017enigma}. The meta-learned models of \citet{dasgupta2020theory} and \citet{binz2020heuristics}, on the other hand, were able to readily show context-specific biases when trained on an appropriate task distribution. 


\subsection{Language Understanding}

Meta-learning may also help us to answer questions regarding how people process, understand, and produce language. Whether the inductive biases needed to acquire a language are learned from experience or are inherited is one of these questions \citep{yang2022one}. \citet{mccoy2020universal} investigated how to equip a model with a set of linguistic inductive biases that are relevant to human cognition. Their solution to this problem builds upon the idea of model-agnostic meta-learning \citep{finn2017model}. In particular, they meta-learned the initial weights of a neural network such that the network can adapt itself quickly to new languages using standard gradient-based learning. When being trained on a distribution over languages, these initial weights can be interpreted as universal factors that are shared across all languages. They showed that this approach identifies inductive biases (e.g. a bias for treating certain phonemes as vowels) that are useful for acquiring a language’s syllable structure. While their work focused on various modeling aspects, they suggested that their framework \say{could $[\ldots]$ be used to empirically investigate the effects that those inductive biases have.} They additionally argued that a Bayesian modeling approach would only be able to consider a restrictive set of inductive biases as it needs to commit to a particular representation and inference algorithm. In contrast, the meta-learning framework made it easy to implement the intended inductive biases by simply manipulating the distribution of encountered languages.

The ability to compose simple elements into complex entities is at the heart of human language. The property of languages to ``make infinite use of finite means'' \citep{chomsky2014aspects} is what allows us to make strong generalizations from limited data. For example, people readily understand what it means to \say{dax twice} or to \say{dax slowly} after learning about the meaning of the verb \say{dax.} How to build models with a similar proficiency, however, remains an open research question. \citet{lake2019compositional} showed that a transformer-like neural network can be trained to make such compositional generalizations through meta-learning. Importantly, during meta-learning, his models were adapted to problems that required compositional generalization, and could thereby acquire the skills needed to solve entirely new problems. \citet{lake2019compositional} argued that meta-learning \say{has implications for understanding how people generalize compositionally.} In particular, it highlights the importance of \say{tackling a series of changing learning problems rather than iterating through a static data-set}, as it is done in traditional neural network training schemes. 

\subsection{Inductive Biases}

Human cognition comes with many useful inductive biases beyond the ability to reason compositionally. The preference for simplicity is one of these biases \citep{chater2003simplicity, feldman2016simplicity}. We readily extract abstract low-dimensional rules that allow us to generalize entirely new situations. Meta-learning is an ideal tool to build models with similar preferences because we can easily generate tasks based on simple rules and use them for meta-learning, thereby enabling an agent to acquire the desired inductive bias from data.




Towards this end, \citet{Kumar2020-gm} tested humans and meta-reinforcement agents on a family of structured tasks generated by a grammar and compared their performance to those trained on a non-structured version of the same task with matched statistics. They expanded these results to a larger suite of tasks generated from simple abstract rules in \citet{kumar2022disentangling}. Humans found it easier to learn in structured tasks, confirming that they have strong priors towards simple abstract rules \citep{Schulz2017-ki}. However, their analysis also indicated that meta-learning is easier on non-structured tasks than on structured tasks. In follow-up work, they found that this result also holds for agents that were trained purely on the structured version of their task but evaluated on both versions \citep{kumar2022using} -- a quite astonishing finding considering that one would expect an agent to perform better on the task distribution it was trained on. The authors addressed this mismatch between humans and meta-learned agents by guiding agents during training to reproduce natural language descriptions that people provided to describe a given task. They found that grounding meta-learned agents in natural language descriptions not only improved their performance but also led to more human-like inductive biases, demonstrating that natural language can serve as a source for abstractions within human cognition.


Their line of work utilizes another interesting technique for training meta-learning agents \citep{kumar2022disentangling, kumar2022using}. It does not rely on a hand-designed task distribution but instead involves sampling tasks from the prior distribution of human participants using a technique known as Gibbs sampling with people \citep{sanborn2007markov, harrison2020gibbs}. While doing so provides them with a data-set of tasks, no expression of the corresponding prior distribution over them is accessible and, hence, it is non-trivial to define a Bayesian model for the given setting. A meta-learned agent, on the other hand, was readily obtained by training on the collected samples.

\subsection{Model-Based Reasoning}

Many realistic scenarios afford two distinct types of learning: model-free and model-based. Model-free learning algorithms directly adjust their strategies using observed outcomes. Model-based learning algorithms, on the other hand, learn about the transition and reward probabilities of an environment, which are then used for downstream reasoning tasks. People are generally thought to be able to perform model-based learning, at least to some extent, and assuming that the problem at hand calls for it \citep{daw2011model, kool2016does}. \citet{wang2016learning} showed that a meta-learned algorithm can display model-based behavior, even if it was trained through a pure model-free reinforcement learning algorithm (see Figure~\ref{fig:examples}c). 

Having a model of the world also acts as the basis for causal reasoning. Traditionally, making causal inferences relies on the notion of Pearl's do-calculus \citep{pearl2009causality}. \citet{dasgupta2019causal}, however, showed that meta-learning can be used to create models that draw causal inferences from observational data, select informative interventions, and make counterfactual predictions. While they have not related their model to human data directly, it could in future work serve as the basis to study how people make causal judgments in complex domains and explain why and when they deviate from normative causal theories \citep{bramley2017formalizing, gerstenberg2021counterfactual}.

Together, these two examples highlight that model-based reasoning capabilities can emerge internally in a meta-learned model if they are beneficial for solving the encountered problem. While there are already many traditional models that can perform such tasks, these models are often slow at run-time as they typically involve Bayesian inference, planning, or both. Meta-learning, on the other hand, \say{shifts most of the compute burden from inference time to training time [which] is advantageous when training time is ample but fast answers are needed at run-time} \citep{dasgupta2019causal}, and may therefore explain how people can perform such intricate computations within a reasonable time-frame.

While model-based reasoning is an emerging property of meta-learned models, it may also be integrated explicitly into such models should it be desired. \citet{jensen2023recurrent} have taken this route, and augmented a standard meta-reinforcement learning agent with the ability to perform temporally extended planning using imagined rollouts. In each time-step, their agent can decide to perform a planning operation instead of directly interacting with the environment (in this case, a spatial navigation task). Their meta-learned agents opted to perform this planning operation consistently after training. Importantly, the model showed striking similarities to patterns of human deliberation by performing more planning early on and with an increased distance to the goal. Furthermore, they found that patterns of hippocampal replays resembled the rollouts of their model.

\subsection{Exploration}

People do not only have to integrate observed information into their existing knowledge, but they also have to actively determine what information to sample. They constantly face situations that require them to decide whether they should explore something new or whether they should rather exploit what they already know. Previous research suggests that people solve this exploration-exploitation dilemma using a combination of directed and random exploration strategies \citep{wilson2014humans, wu2018generalization, gershman2018deconstructing, schulz2019algorithmic}. Why do people use these particular strategies and not others? \citet{binz2022modeling} hypothesized that they do so because human exploration follows resource-rational principles. To test this claim, they devised a family of resource-rational reinforcement learning algorithms by combining ideas from meta-learning and information theory. Their meta-learned model discovered a diverse set of exploration strategies, including random and directed exploration, that captured human exploration better than alternative approaches. In this domain, meta-learning offered a direct path towards investigating the hypothesis that people try to explore optimally but are subject to limited computational resources, whereas designing hand-crafted models for studying the same question would have been more intricate. 

It is not only important to decide how to explore, but also to decide whether exploration is worthwhile in the first place. \citet{lange2020learning} studied this question using the meta-learning framework. Their meta-learned agents are able to flexibly interpolate between implementing exploratory learning behaviors and hard-coded, non-learning strategies. Importantly, which behavior was realized crucially depended on environmental properties, such as the diversity of the task distribution, the task complexity, and the agent's lifetime. They showed, for instance, that agents with a short lifetime should opt for small rewards that are easy to find, while agents with an extended lifetime should spend their time exploring the environment. The study of \citet{lange2020learning} clearly demonstrates that meta-learning makes it conceptually easy to iterate over different environmental assumptions inside a rational analysis of cognition. They only had to modify the environment as desired, followed by rerunning their meta-learning procedure. In contrast, traditional modeling approaches would require hand-designing a new optimal agent each time an environmental change occurs. 

\subsection{Cognitive Control}

Humans are remarkable at adapting to task-specific demands. The processes behind this ability are collectively referred to as cognitive control \citep{botvinick2001conflict}. \citet{cohen2017cognitive} even argues that \say{the capacity for cognitive control is perhaps the most distinguishing characteristic of human behavior.} It should therefore come as no surprise that cognitive control has received a significant amount of attention from a computational perspective \citep{collins2013cognitive, botvinick2014computational}. Recently, some of these computational investigations have been extended to the meta-learning framework. 

The ability to adjust computational resources as needed is one hallmark of cognitive control. \citet{moskovitz2022unified} proposed a meta-learned model with such characteristics. Their model learns a simple default policy -- similar to the model of \citet{binz2022modeling} -- that can be overwritten by a more complex one if necessary. They demonstrate that this model is not only able to capture behavioral phenomena from the cognitive control literature but also known effects in decision-making and reinforcement learning tasks, thereby linking the three domains. Importantly, their study highlights that the meta-learning framework offers the means to account for multiple computational costs instead of just a single one -- in this case, a cost for implementing the default policy and one for deviating from it. 

Taking contextual cues into consideration is another vital aspect of cognitive control. \citet{dubey2020connecting} implemented this idea in the meta-learning framework. In their model, contextual cues determine the initialization of a task-specific neural network that is then trained using model-agnostic meta-learning. They showed that such a model captures \say{the context-sensitivity of human behavior in a simple but well-studied cognitive control task.} Furthermore, they demonstrated that it scales well to more complex domains (including tasks from the MuJoCo \citep{todorov2012mujoco}, CelebA \cite{liu2015faceattributes} and MetaWorld \citep{yu2020meta} benchmarks), thereby opening up new opportunities for modeling human behavior in naturalistic scenarios.

\section{Why Is Not Everything Meta-Learned?}

We have laid out different arguments that make meta-learning a useful tool for constructing cognitive models, but it is important to note that we do not claim that meta-learning is the ultimate solution to every modeling problem. Instead, it is essential to understand when meta-learning is the right tool for the job and when not. 

\subsection{Lack of Interpretability}

Perhaps its most significant detriment is that meta-learning never provides us with analytical solutions that we can inspect, analyze and reason about. In contrast to this, some Bayesian models have analytical solutions. Take as an example the data-generating distribution that we encountered earlier (Equations \ref{eq:priord}-\ref{eq:likelihood}). For these assumptions, a closed-form expression of the posterior predictive distribution is available. By looking at this closed-form expression, researchers have generated new predictions and subsequently tested them empirically \citep{dayan2000explaining, daw2008semi, gershman2015unifying}. Performing the same kind of analysis with a meta-learned model is not as straightforward. We do not have access to an underlying mathematical expression, which makes a structured exploration of theories much harder. 

That being said, there are still ways to analyze a meta-learned model’s behavior. For one, it is possible to use model architectures that facilitate interpretability. \citet{binz2020heuristics} relied on this approach and designed a neural network architecture that produced weights of a probit regression model which were then used to cluster applied strategies into different categories. Doing so enabled them to identify which strategy was used by their meta-learned model in a particular situation. 

Recently, researchers have also started to use tools from cognitive psychology to analyze the behavior of black-box models \citep{ritter2017cognitive, rich2019lessons, schulz2020computational}. For example, it is possible to treat such models just like participants in a psychological experiment and use the collected data to analyze their behavior similar to how psychologists would analyze human behavior \citep{rahwan2019machine, schramowski2022large, binz2022using, dasgupta2022language}. We believe that this approach has great potential for analyzing increasingly capable and opaque artificial agents, including those obtained via meta-learning.

\subsection{Intricate Training Processes}

When using the meta-learning framework, one also has to deal with the fact that training neural networks is complex and takes time. Neural network models contain many moving parts, like weight initializations or the used optimizer, that have to be chosen appropriately such that training can take off in the first place, and training itself may take hours or days until it is finished. When we want to modify assumptions in the data-generating distribution, we have to retrain the whole system from scratch altogether. Thus, although the process of iterating over different environmental assumptions is conceptually straightforward in the meta-learning framework, it may be time-consuming. Bayesian models can, in comparison, sometimes be more quickly adapted to changes in environmental assumptions. To illustrate this, let us assume that you wanted to explain human behavior through a meta-learned model that was trained on the data-generating distribution from Equations \ref{eq:priord}-\ref{eq:likelihood}, but found that the resulting model does not fit the observed data well. Next, you want to consider the alternative hypothesis that people assume a non-stationary environment. While this modification could be done quickly in the corresponding Bayesian model, the meta-learning framework requires retraining on newly generated data.

There is, furthermore, no guarantee that a fully converged meta-learned model actually implements a Bayes-optimal learning algorithm. While we were able to compare to analytical solutions for simple cases like our insect length example, it is in general impossible to verify that a meta-learned algorithm is optimal. Indeed, there are reported cases in which meta-learning failed to find the Bayes-optimal solution \citep{DBLP:journals/corr/abs-2102-02926}. We believe that this issue can be somewhat mitigated by validating meta-learned models in various different ways. But, ultimately future work should come up with techniques to verify meta-learned models. 


\subsection{Meta-Learned or Bayesian Inference?}

In summary, both frameworks –- meta-learning and Bayesian inference -- have their unique strengths and weaknesses. The meta-learning framework does and will not replace Bayesian inference but complement it. It broadens our available toolkit and enables researchers to study questions that were previously out of reach. However, there are certainly situations in which traditional Bayesian inference is a more appropriate modeling choice as we have outlined in this section.

\section{The Role of Neural Networks} \label{sec:biases}  

Most of the points we have discussed so far are agnostic regarding the function approximator implementing the meta-learned algorithm. However, at the same time, we have appealed to neural networks at various points throughout the text. When one looks at prior work, it can also be observed that neural networks are the predominant model class in the meta-learning setting. Why is that the case? In addition to their universality, neural networks offer one big opportunity: they provide a flexible framework for engineering different types of inductive biases into a computational model \citep{goyal2022inductive}. In the following section, we will highlight three examples of how previous work has accomplished this. For each of these examples, we take a concept from psychology, and show how it can be readily accommodated in a meta-learned model.

Perhaps one of the most persuasive idea in cognitive modeling is that of gradient-based learning. It is not only at the heart of one of the most influential models -- the Rescorla-Wagner model \citep{rescorla1972theory, gershman2015unifying} -- but also features prominently in many other theories of human learning, such as connectionist models \citep{rumelhart1988parallel}. Even though the earlier outlined meta-learning procedure relies on gradient-based learning in the outer loop, the resulting inner-loop dynamics must bear no resemblance to gradient descent. However, it is possible to construct meta-learned models whose inner-loop updates rely on gradient-based learning. \citet{finn2017model} proposed a meta-learning technique known as model-agnostic meta-learning that finds optimal initial parameters of a feedforward neural network that is subsequently trained via gradient descent. The idea is that these optimal initial parameters allow the feedforward network to generalize to multiple tasks in a minimal number of gradient steps. While their general setup is similar to the one we discussed, it leads to models that learn via gradient descent instead of models that implement a learning algorithm inside the dynamics of a recurrent neural network. \citet{kirsch2021meta} recently brought these two approaches together into a single model. Their proposed architecture consists of multiple recurrent neural networks that interact with each other. Importantly, they showed that one particular configuration of these networks could implement backpropagation in the forward pass, thereby being able to perform gradient-based learning in a memory-based system.


Exemplar-based models -- like the generalized category model \citep{nosofsky2011generalized} -- are one of the most prominent approaches for modeling how people categorize items into different classes \citep{shepard1987toward, kruschke1990alcove}. They categorize a new instance based on the estimated similarity between that instance and previously seen examples. Recently, meta-learned models with exemplar-based reasoning abilities have been proposed for the task of few-shot classification, in which a classifier must generalize based on a training set containing only a few examples. Matching networks \citep{vinyals2016matching} accomplish this by classifying a new data-point using a similarity-weighted combination of categories in the training set. Importantly, similarity is computed over a learned embedding space, thereby ensuring that the model can scale to high-dimensional stimuli. Follow-up work has taken inspiration from another hugely influential model of human category learning and replaced the exemplar-based mechanism used in matching networks with one based on category prototypes \citep{snell2017prototypical}.

Finally, making inferences using similarities to previous experiences is not only useful for supervised learning but also in the reinforcement learning setting. In the reinforcement learning literature, the ability to store and recollect states or trajectories for later use is studied under the name of episodic memory \citep{lengyel2007hippocampal}. It has been argued that episodic memory could be the key to explaining human performance in naturalistic environments \citep{gershman2017reinforcement}. Episodic memory also plays a crucial role in neuroscience, with studies showing that highly rewarding instances are stored in the hippocampus and made available for recall as and when required \cite{blundell2016model}. \citet{ritter2018been} build upon the neural episodic control idea from \citet{pritzel2017neural} and utilize a differential neural dictionary for episodic recall in the context of meta-learning. Their dictionary stores encodings from previously experienced tasks, which can then be later queried as needed. With this addition, their meta-learned model is able to recall previously discovered policies, retrieve memories using category examples, handle compositional tasks, re-instate memories while traversing the environment, and recover a learning strategy people use in a neuroscience-inspired task.


In summary, human cognition comes with a variety of inductive biases and neural networks provide flexible ways to easily incorporate them into meta-learned models of cognition. We have outlined three such examples in the section, demonstrating how to integrate gradient-based learning, exemplar- and prototype-based reasoning, and episodic memory into a meta-learned model. There are, furthermore, many other inductive biases for neural network architectures that could be utilized in the context of meta-learning but have not been yet. Examples include networks that perform differentiable planning \citep{tamar2016value, farquhar2017treeqn}, extract object-based representations \citep{piloto2022intuitive, sancaktar2022curious}, or modify their own connections through synaptic plasticity \citep{schlag2021linear, miconi2020backpropamine}.

\section{Towards a Domain-General Model of Human Learning}

What does the future hold for meta-learning? The current generation of meta-learned models of cognition is almost exclusively trained on the data-generating distribution of a specific problem family. While this training process enables them to generalize to new tasks inside this problem family, they are unlikely to generalize to completely different domains. We would, for example, not expect a meta-learned algorithm to perform a challenging maze navigation task if it was only trained to predict the lengths of insect species.

While domain-specific models have (and will continue to) provide answers to important research questions, we agree with \citet{newell1992unified} that \say{unified theories of cognition are the only way to bring this wonderful, increasing fund of knowledge under intellectual control.} Ideally, such a unified theory should manifest itself in a domain-general cognitive model that cannot only solve prediction tasks but is also capable of human-like decision-making \citep{gigerenzer2011heuristic}, category learning \citep{ashby2005human}, navigation \citep{montello2005navigation}, problem-solving \citep{newell1972human} and so on. We consider the meta-learning framework the ideal tool for accomplishing this goal as it allows us to compile arbitrary assumptions about an agent's beliefs of the world (arguments 1 and 2) and its computational architecture (arguments 3 and 4) into a cognitive model.

To obtain such a domain-general cognitive model via meta-learning, however, a few challenges need to be tackled. First of all, there is the looming question of how a data-generating distribution that contains many different problems should be constructed. Here, we may take inspiration from the machine learning community, where researchers have devised generalist agents by training neural networks on a large set of problems \citep{reed2022generalist}. \citep{team2023human} have recently shown that this is a promising path for scaling up meta-learning models. They trained a meta-reinforcement learning agent on a vast open-ended world with over $10^{40}$ possible tasks. The resulting agent can adapt to held-out problems as quickly as humans, and \say{displays on-the-fly hypothesis-driven exploration, efficient exploitation of acquired knowledge, and can successfully be prompted with first-person demonstrations.} In the same vein, we may come up with a large collection of tasks that are more commonly used to study human behavior \citep{yang2019task, molano2022neurogym, miconi2023large}, and use them to train a meta-learned model of cognition.


Language will likely play an important role in future meta-learning systems. We do not want a system that learns every task from scratch via trial and error but one that can be provided with a set of instructions similar to how a human subject would be instructed in a psychological experiment. Having agents capable of language will not only enable them to understand new tasks in a zero-shot manner but may also facilitate their cognitive abilities. It, for example, allows them to decompose tasks into sub-tasks, learn from other agents, or generate explanations \citep{colas2022language}. Fortunately, our current best language models \citep{brown2020language, chowdhery2022palm} and meta-learning systems are both based on neural networks. Thus, integrating language capabilities into a meta-learned model of cognition should -- at least conceptually -- be fairly straightforward. Doing so would furthermore enable such models to harvest the compositional nature of language to make strong generalizations to tasks outside of the meta-learning distribution. The potential for this was highlighted in a recent study of \citep{riveland2022generalization} which found that language-conditioned recurrent neural network models can perform entirely novel psychophysical tasks with high accuracy. 

Moreover, a sufficiently general model of human cognition must not only be able to select amongst several given options, like in a decision-making or category learning setting, but it also needs to maneuver within a complex world. For this, it needs to perceive and process high-dimensional visual stimuli, it needs to control a body with many degrees of freedom, and it needs to actively engage with other agents. Many of these problems have been at the heart of the deep learning community \citep{Hill2020Environmental, mcclelland2020placing, team2021open, strouse2021collaborating}, and it will be interesting to see whether the solutions developed there can be integrated into a meta-learned model of cognition.

Finally, there are also some challenges on the algorithmic side that need to be taken into account. In particular, it is unclear how far currently used model architectures and outer-loop learning algorithms scale. While contemporary meta-learning algorithms are able to find approximately Bayes-optimal solutions to simple problems, they sometimes struggle to do so on more complex ones (e.g. as in the earlier discussed work of \citet{DBLP:journals/corr/abs-2102-02926}). Therefore, it seems likely that simply increasing the complexity of the meta-learning distribution will not be sufficient -- we will also need model architectures and outer-loop learning algorithms that can handle increasingly complex data-generating distributions. The transformer architecture \citep{vaswani2017attention}, which has been very successful at training large language models \citep{brown2020language, chowdhery2022palm}, provides one promising candidate, but there could be countless other (so far undiscovered) alternatives. 

\begin{figure}[!t]
\centering
\hspace*{-0.25cm}\includegraphics[width=1.03\textwidth]{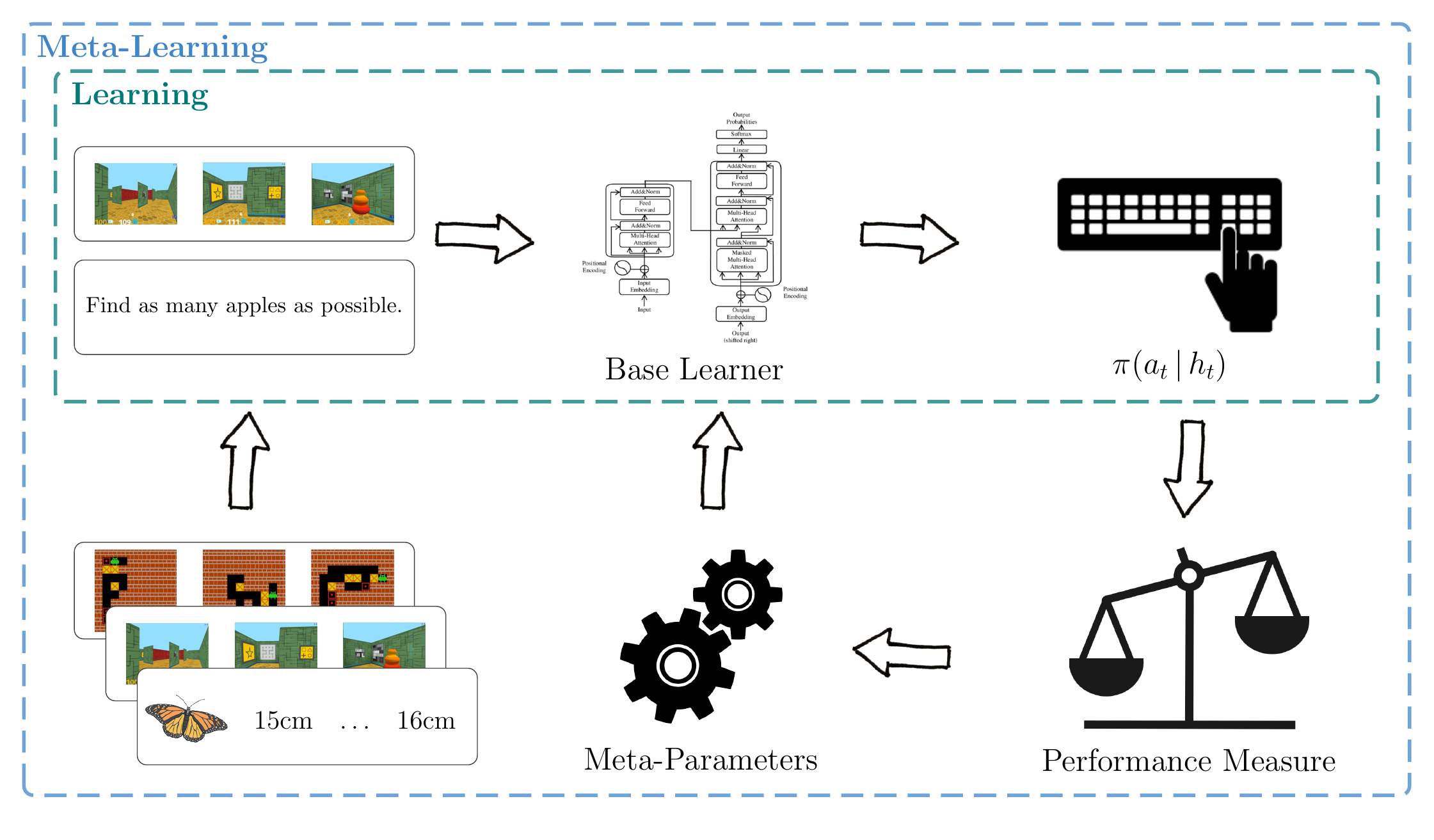}
\caption{Illustration of how a domain-general meta-learned model of cognition could look like. Modifications include training on more diverse task distributions, providing natural language instructions as additional inputs, and relying on scalable model architectures.}
\label{fig:domaingeneral}
\end{figure}

Thus, taken together, there are still substantial challenges involved in creating a domain-general meta-learned model of cognition. In particular, we have argued in this section that we need to (1) meta-learn on more diverse task distributions, (2) develop agents that can parse instructions in form of natural language, (3) embody these agents in realistic problem settings, and (4) find model architectures that scale to these complex problems. Figure \ref{fig:domaingeneral} summarizes these points graphically.

\section{Conclusion}

Most computational models of human learning are hand-designed, meaning that at some point a researcher sat down and defined how they behave. Meta-learning starts with an entirely different premise. Instead of designing learning algorithms by hand, one trains a system to achieve its goals by repeatedly letting it interact with an environment. While this seems quite different from traditional models of learning on the surface, we have highlighted that the meta-learning framework actually has a deep connection to the idea of Bayesian inference, and thereby to the rational analysis of cognition. Using this connection as a starting point, we have highlighted several advantages of the meta-learning framework for constructing rational models of cognition. Together, our arguments demonstrate that meta-learning cannot only be applied in situations where Bayesian inference is impossible but also facilitates the inclusion of computational constraints and neuroscientific insights into rational models of human cognition. Earlier criticisms of the rational analysis of cognition have repeatedly pointed out that \say{rational Bayesian models are significantly unconstrained} and that they should be \say{developed in conjunction with mechanistic considerations to offer substantive explanations of cognition} \citep{jones2011bayesian}. We believe that the meta-learning framework provides the ideal opportunity to do so as it allows for a painless integration of flexible computational mechanisms.

It is worth pointing out that meta-learning can be also motivated by taking neural networks as a starting point. From this perspective, it bridges two of the most popular theories of cognition -- Bayesian models and connectionism -- by bringing the scalability of neural network models into the rational analysis of cognition. We, therefore, believe that meta-learning provides a powerful tool to scale up psychological theories to more complex settings. However, at the same time, meta-learning has not delivered on this promise yet. Existing meta-learned models of cognition are typically applied to classical scenarios where established models already exist. Thus, we have to ask: what prevents the application to more complex and general paradigms? First, such paradigms themselves have to be developed. Fortunately, there is currently a trend toward measuring human behavior on more naturalistic tasks \citep{brandle2022intrinsically, brandle2022exploration, schulz2019structured}, and it will be interesting to see what role meta-learning will play in modeling behavior in such settings. Furthermore, meta-learning can be intricate and time-consuming. We hope that the present article -- together with the accompanying code examples -- makes this technique less opaque and more accessible to a wider audience. Future advances in hardware will likely make the meta-learning process quicker and we are therefore hopeful that meta-learning can ultimately fulfill its promise of identifying plausible models of human cognition in situations that are out of reach for hand-designed algorithms.

\begin{figure}
\begin{theo} \normalfont \small
We proof that the posterior predictive distribution $p(x_{t+1} | x_{1:t})$ maximizes the log-likelihood of future observations averaged over the data-generating distribution:
\begin{equation}
    p(x_{t+1} | x_{1:t}) = \argmax_q \mathbb{E}_{p(\mu, x_{1:t+1})} \left[ \log q(x_{t+1} | x_{1:t})\right]
\end{equation}

The essence of this proof is to show that the posterior predictive distribution is superior to any other reference distribution $r(x_{t+1} | x_{1:t})$ in terms of log-likelihood:
\begin{align*}
    \mathbb{E}_{p(\mu, x_{1:t})} \left[ 
	\log p(x_{t+1}  | x_{1:t}) \right] \geq \mathbb{E}_{p(\mu, x_{1:t})} \left[ 
	\log r(x_{t+1} | x_{1:t}) \right] 
\end{align*}

or equivalently that:
\begin{align*}
    \mathbb{E}_{p(\mu, x_{1:t})} \left [ \log \frac{p(x_{t+1}  | x_{1:t})}{r(x_{t+1}  | x_{1:t})} \right] \geq 0
\end{align*}

Proofing this conjecture is straight-forward \citep{aitchison1975goodness}:
\begin{flalign*}
    &\mathbb{E}_{p(\mu, x_{1:t})} \left [ \log \frac{p(x_{t+1}  | x_{1:t})}{r(x_{t+1}  | x_{1:t})} \right] \\
	= &\sum_{\mu} \sum_{x_{1:t}} \sum_{x_{t+1}} \log \frac{p(x_{t+1} | x_{1:t})}{r(x_{t+1} | x_{1:t})} p(x_{t+1} | \mu )p(x_{1:t} | \mu) p(\mu) & \hspace{0.5em} & \text{\setstretch{1.0}\color{mpigreen}\footnotesize\makecell[l]{definition of \\  expectation}} \\
	= &\sum_{x_{1:t}} \sum_{\mu} \sum_{x_{t+1}} \log \frac{p(x_{t+1} | x_{1:t})}{r(x_{t+1} | x_{1:t})} p(x_{t+1} | \mu )p(x_{1:t} | \mu) p(\mu)  & \hspace{0.5em} &  \text{\setstretch{1.0}\color{mpigreen}\footnotesize\makecell[l]{change order \\ of summation}}\\
	= &\sum_{x_{1:t}} \sum_{\mu} \sum_{x_{t+1}} \log \frac{p(x_{t+1} | x_{1:t})}{r(x_{t+1} | x_{1:t})} p(x_{t+1} | \mu )p(\mu | x_{1:t}) p(x_{1:t})    & \hspace{0.5em} &  \text{\color{mpigreen}\footnotesize Bayes' rule}\\
	= &\sum_{x_{1:t}} \left[ \sum_{\mu} \sum_{x_{t+1}} \log \frac{p(x_{t+1} | x_{1:t})}{r(x_{t+1} | x_{1:t})} p(x_{t+1} | \mu )p(\mu | x_{1:t}) \right] p(x_{1:t})    & \hspace{0.5em} &  \text{\color{mpigreen}\footnotesize\makecell[l]{factor out \\ $p(x_{1:t})$}}\\
	= &\sum_{x_{1:t}} \left[ \sum_{x_{t+1}} \sum_{\mu} \log \frac{p(x_{t+1} | x_{1:t})}{r(x_{t+1} | x_{1:t})} p(x_{t+1} | \mu )p(\mu | x_{1:t}) \right] p(x_{1:t})     & \hspace{0.5em} &  \text{\setstretch{1.0}\color{mpigreen}\footnotesize\makecell[l]{change order \\ of summation}}\\
	= &\sum_{x_{1:t}} \left[ \sum_{x_{t+1}} \log \frac{p(x_{t+1} | x_{1:t})}{r(x_{t+1} | x_{1:t})} \left[\sum_{\mu} p(x_{t+1} | \mu )p(\mu | x_{1:t})  \right] \right] p(x_{1:t})    & \hspace{0.5em} &  \text{\setstretch{1.0}\color{mpigreen}\footnotesize\makecell[l]{factor out \\ log-term}}\\
	= &\sum_{x_{1:t}} \left[ \sum_{x_{t+1}} \log \frac{p(x_{t+1} | x_{1:t})}{r(x_{t+1} | x_{1:t})} p(x_{t+1} | x_{1:t})  \right] p(x_{1:t})   & \hspace{0.5em} &  \text{\setstretch{1.0}\color{mpigreen}\footnotesize Equation \ref{eq:ppd}}\\
	= &\sum_{x_{1:t}} \text{KL} \left[p(x_{t+1} | x_{1:t}) || r(x_{t+1} | x_{1:t}) \right] p(x_{1:t})   & \hfill & \text{\setstretch{1.0}\color{mpigreen}\footnotesize\makecell[l]{definition of \\  KL divergence}} \\
	&\geq 0 
     & \hfill & {\color{mpigreen}\blacksquare}
\end{flalign*}

Note that while we used sums in our proof, thereby assuming that relevant quantities take discrete values, the same ideas can be readily applied to continuous-valued quantities by replacing sums with integrals.
\end{theo}
    \label{fig:my_label}
\end{figure}

\begin{figure}
\begin{theo} \normalfont \small
The main text has focused on tasks in which an agent receives direct feedback about which response would have been correct. In the real world, however, people do not always receive such explicit feedback. They, instead, often have to deal with partial information -- taking the form of rewards, utilities, or costs -- that merely informs them about the quality of their response. \\

Problems that fall into this category are often modeled as Markov decision processes (MDPs). In an MDP, an agent repeatedly interacts with an environment. In each time-step, it observes the state of the environment $s_t$ and can take an action $a_t$ that leads to a reward signal $r_t$ sampled from a reward distribution $p(r_t | s_t, a_t, \mu_r)$. Executing an action furthermore influences the environment state at the next time-step according to a transition distribution $p(s_{t+1} | s_{t}, a_t, \mu_s)$.  \\


The goal of a Bayes-optimal reinforcement learning agent is to find a policy, which is a mapping from a history of observations $h_t = s_1, a_1, r_1, \ldots, s_{t-1}, a_{t-1}, r_{t-1}, s_t$ to a probability distribution over actions $\pi^*(a_t | h_t)$, that maximizes the total amount of obtained rewards across a finite horizon $H$ averaged over all problems that may be encountered:
\begin{equation} \label{eq:rlobj} 
 \pi^*(a_t | h_t) = \argmax_{\pi}~ \mathbb{E}_{p(\mu_r, \mu_s)\prod p(r_t | s_t, a_t, \mu_r)p(s_{t+1} | s_{t}, a_t, \mu_s)\pi(a_t | h_{t})} \left[ \sum_{t=1}^H r_t \right]
\end{equation}

MDPs with unknown reward and transition distributions are substantially more challenging to solve optimally compared to supervised problems as there is no teacher informing the agent about which actions are right or wrong. Instead, the agent has to figure out the most rewarding course of action solely through trial and error. Finding an analytical solution to Equation \ref{eq:rlobj} is extremely challenging and indeed only possible for a few special cases \citep{gittins1979bandit, duff2003optimal}, which made it historically near impossible to investigate such problems within the framework of rational analysis. \\

Even though finding an analytical expression of the Bayes-optimal policy is often impossible, it is straightforward to meta-learn an approximation to it \citep{duan2016rl, wang2016learning}. The general concept is almost identical to the supervised learning case: parameterize the to-be-learned policy with a recurrent neural network and replace the maximization over the set of all possible policies from Equation \ref{eq:rlobj} with a sample-based approximation that maximizes over neural network parameters. The resulting problem can then be solved using any standard deep reinforcement learning algorithm.  \\

Like in the supervised learning case, the resulting recurrent neural network implements a free-standing reinforcement learning algorithm after meta-learning is completed. Learning is once again implemented via a simple forward pass through the network, i.e., by conditioning the model on an additional data-point. The meta-learned reinforcement learning algorithm approximates the Bayes-optimal policy under the same conditions as in the supervised learning case: a sufficiently expressive model and an optimization procedure that is able to find the global optimum. 

\end{theo}
    \label{fig:box_rl}
\end{figure}

\newpage 

\subsection*{Acknowledgements}

\noindent The authors would like to thank Sreejan Kumar, Tobias Ludwig, Dominik Endres, and Adam Santoro for their valuable feedback on an earlier draft.
  
\subsection*{Funding statement}

\noindent This work was funded by the Max Planck Society, the Volkswagen Foundation, as well as the Deutsche Forschungsgemeinschaft (DFG, German Research Foundation) under Germany’s Excellence Strategy–EXC2064/1–390727645. 

\subsection*{Conflicts of Interest statement}

\noindent The author(s) declare none.

\bibliography{bibliography}

\end{document}